\documentclass[a4paper,UKenglish,cleveref, autoref, thm-restate]{lipics-v2021}
\usepackage{tikz}
\usepackage{graphicx} % Required for inserting images
\usepackage{booktabs}
\usepackage{amsmath}
\usetikzlibrary{arrows.meta,positioning,patterns}
\usepackage[dvipsnames]{xcolor}
\usepackage[ruled,vlined,linesnumbered,algo2e]{algorithm2e}

\usepackage{xcolor}

\definecolor{jobone}{RGB}{49,132,253}   % J1
\definecolor{jobtwo}{RGB}{251,90,90}    % J2
\definecolor{jobthree}{RGB}{40,201,120} % J3
\newcommand{\lanesep}{1}
\tikzset{
  opJ1/.style={draw=black,fill=jobone},
  opJ2/.style={draw=black,fill=jobtwo},
  opJ3/.style={draw=black,fill=jobthree},
  predJ1/.style={-Latex,very thick,jobone},
  predJ2/.style={-Latex,very thick,jobtwo},
  predJ3/.style={-Latex,very thick,jobthree},
}

\newcommand{\Op}[5]{%
  \pgfmathsetmacro{\y}{-(#1-1)*\lanesep}
  \pgfmathtruncatemacro{\prod}{(#4-1)*3+#5}
  \draw[opJ#4] (-1.8+#2*0.5,\y-0.4) rectangle ++(#3*0.5,0.6);
  \node[font=\scriptsize] at (-1.8+#2*0.5+0.5*#3*0.5,\y - 0.1) {$o_{\prod}$};
  \coordinate (o_{#4, #5}) at (-1.8+#2*0.5+0.25*#3,\y);
}

\newcommand{\Optv}[5]{%
  \pgfmathsetmacro{\y}{-(#1-1)*\lanesep}
  \pgfmathtruncatemacro{\prod}{(#4-1)*3+#5}
  \draw[opJ#4] (-1.8+#2*0.5,\y-0.4) rectangle ++(#3*0.5,0.6);
  \node[font=\scriptsize, rotate=90] at (-1.8+#2*0.5+0.5*#3*0.5,\y - 0.1) {$o_{\prod}$};
  \coordinate (o_{#4, #5}) at (-1.8+#2*0.5+0.25*#3,\y);
}
\newcommand{\Opv}[5]{%
  \pgfmathsetmacro{\y}{-(#1-1)*\lanesep}
  \pgfmathtruncatemacro{\prod}{(#4-1)*3+#5}
  \draw[opJ#4, fill opacity=0.3, postaction={pattern=north east lines, pattern color=black!30}] (-1.8+#2*0.5,\y-0.4) rectangle ++(#3*0.5,0.6);
  \node[font=\scriptsize] at (-1.8+#2*0.5+0.5*#3*0.5,\y - 0.1) {$o_{\prod}$};
  \coordinate (o_{#4, #5}) at (-1.8+#2*0.5+0.25*#3,\y);
}
\newcommand{\Ops}[7]{%
  \pgfmathsetmacro{\y}{-(#1-1)*\lanesep}
  \pgfmathtruncatemacro{\prod}{(#4-1)*3+#5}
  \draw[opJ#4, fill opacity=0.3, postaction={pattern=north east lines, pattern color=black!30}] (-1.8+#2*0.5,\y-0.4+0.6*#6/#7) rectangle ++(#3*0.5,0.6/#7);
  \node[font=\scriptsize] at (-1.8+#2*0.5+0.5*#3*0.5,\y - 0.1 +#6*0.6/#7 - 1*0.2) {$o_{\prod}$};
  \coordinate (o_{#4, #5}) at (-1.8+#2*0.5+0.25*#3,\y);
}

\title{CP or DP? Why Not Both: A Case Study in the Partial Shop Scheduling Problem.}

\titlerunning{A Case Study in the Partial Shop Scheduling Problem}

\newcommand{\uclouvainaffil}{ICTEAM, UCLouvain, Belgium}

\author{Emma Legrand}{\uclouvainaffil}{emma.legrand@uclouvain.be}{https://orcid.org/0009-0000-3836-1782}{}

\author{Roger Kameugne}{\uclouvainaffil}{roger.kameugne@uclouvain.be}{https://orcid.org/0000-0003-1809-9822}{}

\author{Pierre Schaus}{\uclouvainaffil}{pierre.schaus@uclouvain.be}{https://orcid.org/0000-0002-3153-8941}{}

\authorrunning{E. Legrand, R. Kameugne, and P. Schaus}

\Copyright{Emma Legrand, Roger Kameugne, and Pierre Schaus}

\ccsdesc[100]{Mathematics of computing~Combinatorial optimization} 

\keywords{Partial Shop Scheduling Problem, dynamic programming, decision diagram, constraint programming, Large Neighborhood Search} 

\category{} 

\relatedversion{} 
\nolinenumbers %uncomment to disable line numbering

\supplementdetails[subcategory={Source Code}, swhid={...}]{Software}{https://anonymous.4open.science/r/DP-CP\_JobShop-8C5E}

\EventEditors{John Q. Open and Joan R. Access}
\EventNoEds{2}
\EventLongTitle{42nd Conference on Very Important Topics (CVIT 2016)}
\EventShortTitle{CVIT 2016}
\EventAcronym{CVIT}
\EventYear{2016}
\EventDate{December 24--27, 2016}
\EventLocation{Little Whinging, United Kingdom}
\EventLogo{}
\SeriesVolume{42}
\ArticleNo{23}
\begin{document}

\maketitle

\begin{abstract}

Dynamic Programming (DP) and Constraint Programming (CP) are well-established paradigms for solving combinatorial optimization problems. Usually, these two approaches are used separately. This paper aims to show that the two can be combined effectively and elegantly, with DP serving as the primary search framework and CP used as a subroutine to leverage global constraint propagation. This paper presents such an approach for the Partial Shop Scheduling Problem (PSSP), for which a pure DP method has previously been proposed, and efficient CP filtering algorithms are available. The PSSP is a general scheduling problem where each job consists of a set of operations with arbitrary precedence constraints. The approach is flexible enough to accommodate anytime DP strategies, such as anytime column search, whereas the original DP algorithm operated in a strictly layer-wise manner.
Moreover, the flexibility of the CP modeling makes it straightforward to incorporate arbitrary precedence constraints. As a result, the model naturally handles any precedence graph and even enables the design of a Large Neighborhood Search (LNS) scheme, in which the DP model is reused, and partial-order schedules are imposed across restarts to improve the incumbent solution.
While not competitive with state-of-the-art pure CP solvers for this specific problem, our primary contribution is demonstrating the viability of this hybrid integration.

\end{abstract}

\section{Introduction}
This work considers solving the \emph{Partial Shop Scheduling Problem} (PSSP) using dynamic programming (DP), constraint programming (CP), and a hybrid DP-CP approach. The PSSP is a general scheduling problem where each job consists of a set of operations with arbitrary precedence constraints. In \cite{HookerH18}, the authors review the integration of CP and operations research (OR) to solve combinatorial optimization problems. The PSSP generalizes classical scheduling problems such as the job-shop scheduling problem (JSP) and the open-shop scheduling problem (OSP), which are well-studied combinatorial optimization problems across operations research, artificial intelligence, and management science \cite{XiongSRH22,DauzerePeresDST24,pinedo2016scheduling,blazewicz2001scheduling,zhang2019review}. Their enduring relevance stems from the fact that they serve as relaxations or generalizations for numerous scheduling challenges found in a wide range of real-world industrial settings. In these problems, a set of operations must be scheduled on distinct machines, subject to given precedence constraints. The common objective is to minimize the makespan, defined as the maximum completion time of all operations.

These scheduling problems are typically strongly NP-hard \cite{GareyJS76,GonzalezS76} and have been well investigated in the literature \cite{XiongSRH22,DauzerePeresDST24,GromichoHST12,HoornNOG17,Ozolins19,Ozolins20,Ozolins21,zhang2019review}. Despite extensive research into exact methods for these problems, DP approaches remain relatively rare.

For the JSP, Gromicho et al. \cite{GromichoHST12} introduced a DP formulation that views a solution as a sequence of operations, applying specific techniques to maintain optimality guarantees. By leveraging dominance properties to prune the search space, this algorithm successfully solves moderate-sized instances. Hoorn et al. \cite{HoornNOG17} later refined this approach, proving its capability to find optimal solutions for particularly hard instances from established benchmarks \cite{vanHoorn}. Building on this DP formulation, Ozolins \cite{Ozolins20} proposed an improved bounded variant, combining DP with branch-and-bound, experimented on medium-sized instances.

Pure CP approach combines the well-known \textsc{NoOverlap} global constraint \cite{vilim2007global} with specific search strategies \cite{baptiste2001constraint,LaborieRSV18}. To achieve maximal domain filtering at each node of the search tree, this global constraint integrates several propagation techniques: \emph{overload checking}, \emph{edge-finding} (also referred to as \emph{heads and tails adjustment} \cite{carlier1994adjustment} or immediate selection \cite{brucker1994job}), \emph{detectable precedences}, and \emph{not-first/not-last} rules. CP solvers can be employed to iteratively improve rapidly the current solution by relaxing a fraction of the assignments within a Large Neighborhood Search (LNS) framework and then prove optimality with a \emph{failure directed search} \cite{vilim2015failure}.

In this paper, we demonstrate how a CP model can be elegantly integrated as a subroutine within a DP approach to leverage the powerful global constraint propagation implemented in existing CP solvers. To achieve this, the current DP state, the upper bound on the makespan, and the current decision are imposed as constraints within the CP model. Computing the fixpoint of this propagation results in tighter execution intervals for the tasks and a strengthened lower bound for the makespan ultimately reducing the number of transitions in the DP approach. Furthermore, CP can be used as a proxy model to further tighten the lower bound using a dichotomic search based on feasibility checks in each state.

This hybrid DP-CP framework naturally models the PSSP and facilitates a Large Neighborhood Search (LNS) scheme by relaxing a subset of precedences. Empirical evaluations on standard benchmarks confirm the viability of this hybridization, demonstrating that our approach effectively exploits CP global constraints and is competitive with standard pure CP approaches in minimizing the number of explored nodes.
Our contributions are summarized as follows: 
\begin{itemize}
    \item To improve anytime performance, we replace the layer-based DP search proposed in \cite{GromichoHST12} with an \emph{Anytime Column Search} (ACS) \cite{vadlamudi2012anytime}.
    \item We propose a hybridization of DP and CP models where the power of the global constraint \textsc{NoOverlap} plays a central role.  
    \item The new framework natively handles the PSSP and demonstrates its ability to improve solutions in a large neighborhood search.
    \item Empirical evaluations are conducted on well-known suites of benchmarks.
\end{itemize}

The rest of the paper is organized as follows.
\Cref{sec:related} reviews the related work.
\Cref{sec:preli} presents the preliminary concepts used throughout this work.
\Cref{sec:dp} describes the state-of-the-art DP model while \Cref{sec:acs} presents our adaptation of the anytime column search for the layer-based DP.
\Cref{sec:hybrid} details the proposed DP-CP hybrid framework for the PSSP.
The Large Neighborhood Search (LNS) is described in \cref{sec:lns}.
\Cref{sec:xps} provides a comprehensive empirical evaluation of the proposed approach.
Finally, \cref{sec:concl} concludes the paper.

\section{Related Work}\label{sec:related}
A closely related work by Marijnissen et al. \cite{marijnissen2026domain} developped in parallel and independently to this one also proposes a generic framework for integrating CP propagation into Domain-Independent Dynamic Programming (DIDP) \cite{kuroiwa2023domain}. Their work evaluates the synergy between DP and CP on problems such as single-machine scheduling with time windows, RCPSP, and TSPTW, primarily using A* and Complete Anytime Beam Search (CABS).
The DIDP framework of \cite{marijnissen2026domain} aims for domain-independence and generality across problem classes.
It treats CP as a black box for feasibility checks and dual bound computations but does not leverate the filtering power of CP of the domains. Our integration is more problem-specific and deliberately tailored to the structure of the Partial Shop Scheduling Problem (PSSP). We exploit CP propagation in a more deeply coupled way: the \textsc{NoOverlap} global constraint fixpoint is used to \emph{discover new precedence constraints} at each DP transition, and these precedences are explicitly incorporated into the DP state itself. A more minor difference is that we employ Anytime Column Search (ACS) as our primary search strategy, whereas \cite{marijnissen2026domain} focuses on A* and CABS.

\section{The Partial Shop Scheduling Problem (PSSP)}\label{sec:preli}
We consider the \emph{Partial Shop Scheduling Problem} (PSSP) defined by a finite set of operations $O = \{o_1, \dots, o_N\}$ to be executed on a finite set of machines $M=\{M_1, \dots, M_m\}$ with $|M|=m$. This set of operations is partitioned into $n = N/m$ subsets, such that each machine is assigned to exactly one operation in each partition.

Each operation $o \in O$ has a specified processing time $p_o > 0$, requires a specific machine $m(o) \in M$ for its exclusive use without preemption, and is associated with a specific partition $j(o)$.
Furthermore, the operations are subject to a set of precedence constraints represented by a directed acyclic graph (DAG) $G = (O, E)$. An edge $(o, o') \in E$ indicates that operation $o$ must be completed before operation $o'$ can start, denoted $o \lessdot o'$. The set of successors of an operation $o$ is denoted by $\mathit{succs}(o)$ and its predecessors by $\mathit{preds}(o)$.

A solution to this problem is an assignment of start times $\psi_o$ to each operation $o \in O$ such that all precedence constraints ($\psi_{o'} \geq \psi_o + p_o$ for all $(o, o') \in E$) and machine and partition disjunctive constraints (no two operations on the same machine/partition overlap in time) are satisfied.
The objective is to minimize the makespan $C_{\max} = \max_{o \in O} (\psi_o + p_o)$.

This general PSSP formulation encompasses classical problems such as the Job-Shop Scheduling Problem (JSP) and the Open-Shop Scheduling Problem (OSP). In the classical JSP, the set of operations is partitioned into disjoint subsets called jobs, and the precedence graph $E$ consists of a set of disjoint directed paths, one for each job. In the OSP, there are no initial precedence constraints among operations ($E = \emptyset$), but operations belonging to the same job cannot overlap in time, requiring additional disjunctive constraints.

For the JSP instance of three jobs and three machines shown in \cref{tab:instance}, the precedences $E$ is equal to $\left\{(o_1, o_2), (o_2, o_3), (o_4, o_5), (o_5, o_6), (o_7, o_8), (o_8, o_9)\right\}.$ The job 1 is defined by $\{o_1,o_2,o_3\}$, the job 2 is $\{o_4,o_5,o_6\}$ and the job 3 is $\{o_7, o_8, o_9\}$. \cref{tab:solution} gives a solution for the JSP instance of \cref{tab:instance}.
\begin{table}[b]
  \centering
  \begin{minipage}[c]{0.35\textwidth}
    \centering
    \begin{tabular}{c|cc|cc|cc}
        & m& p &m &p &m &p \\
    \midrule
      Job 1 & 1 & 3 & 2 & 2 & 3 & 2 \\
      \midrule
      Job 2 & 1 & 2 & 3 & 1 & 2 & 4 \\
      \midrule
      Job 3 & 2 & 4 & 3 & 3 & 1 & 1
    \end{tabular}
    \caption{Example of a JSP instance with 3 jobs and 3 machines.}
    \label{tab:instance}
  \end{minipage}\hfill
  \begin{minipage}[c]{0.55\textwidth}
    \centering
    \begin{tikzpicture}

    \coordinate (origin) at (0,0);
    
    \begin{scope}[shift={(origin)}]
      % ======= BACKDROP: time grid & axes =======
      \foreach \x in {0,...,11} {
        \draw[very thin,gray!40] (-1.8+\x*0.5,0.2) -- (-1.8+\x*0.5,-1.5*\lanesep-0.9);
      }
      \foreach \x in {0,...,11} {
        \node[font=\scriptsize] at (-1.8+\x*0.5,-2.7) {\x};
      }
      \node[anchor=west] at (-2.6,0) {\large M1};
      \node[anchor=west] at (-2.6,-\lanesep) {\large M2};
      \node[anchor=west] at (-2.6,-2*\lanesep) {\large M3};

      \draw[thick] (-1.8,0.2) rectangle (3.7,-1.5*\lanesep-0.9);

      % Operations
        \Op{1}{0}{2}{2}{1}
      \Op{1}{2}{3}{1}{1}
      \Optv{1}{7}{1}{3}{3}

      \Op{2}{0}{4}{3}{1}
      \Op{2}{5}{2}{1}{2}
      \Op{2}{7}{4}{2}{3}

      \Optv{3}{2}{1}{2}{2}
      \Op{3}{4}{3}{3}{2}
      \Op{3}{7}{2}{1}{3}
    \end{scope}

      % Legend
      \begin{scope}[shift={(origin)}]
        \draw[opJ1] (-2.5,-3.5) rectangle +(1,0.5);\node[anchor=west] at (-1.5,-3.25) {Job 1};
        \draw[opJ2] (-0.3,-3.5) rectangle +(1,0.5);\node[anchor=west] at (0.7,-3.25) {Job 2};
        \draw[opJ3] (1.8,-3.5) rectangle +(1,0.5);\node[anchor=west] at (2.8,-3.25) {Job 3};
      \end{scope}
    \end{tikzpicture}
    \caption{Solution for the instance of Table \ref{tab:instance}}
    \label{tab:solution}
  \end{minipage}
\end{table}

\section{Dynamic Programming Approach}\label{sec:dp}

A DP formulation for the JSP was originally proposed by Gromicho et al. \cite{GromichoHST12}, subsequently refined by Hoorn et al. \cite{HoornNOG17}, and reused by Ozolins \cite{Ozolins20}. In this approach, operations are scheduled by fixing their start times. An operation becomes eligible for scheduling only when all of its predecessors in the precedence graph have already been scheduled. This operation is then scheduled at its earliest possible start time, ensuring no temporal overlap on its required machine. A node in the search space represents a DP state, defined by the earliest possible start times of the remaining unscheduled operations. Because different sequences of scheduling decisions can lead to the exact same state, the search space forms a graph rather than a simple tree.
The transition cost between two nodes corresponds to the incremental increase in the makespan of the partial solution. Hoorn et al. \cite{HoornNOG17} explore this state graph layer by layer, where the shortest path through the graph yields the schedule with the optimal makespan.
Formally the DP state is a triplet $S = \langle \psi, \Lambda, \mathit{done} \rangle$, where:

\begin{itemize}
    \item $\psi$ tracks the earliest possible start time for every operation.
    \item $\Lambda$ records the machine ID of the last scheduled operation.
    \item $\mathit{done}$ represents the set of operations that have already been scheduled.
\end{itemize}

The state $S$ defines a partial sequence, $T_S$, with a current makespan denoted by $C_{\max}(S)$. Based on our eligibility rules, the set of currently available operations is mathematically specified as: $\varepsilon(S)=\{o \in O \setminus \mathit{done}(S) \mid \mathit{preds}(o) \subseteq \mathit{done}(S)\}$

To mitigate the exponential size of the search space, the authors introduced dominance rules. These rules allow the algorithm to safely discard dominated transitions and nodes while mathematically guaranteeing that at least one path to the optimal solution is preserved.

\subsection{Transition Dominance}\label{subsec:trans_dominance}

Given a complete solution and a start-time assignment, this solution can be obtained by successively selecting an operation from $\varepsilon(S)$ and scheduling it as early as possible without violating machine constraints, continuing until all tasks are scheduled. Because multiple tasks may be eligible at a given stage, different task selection sequences can yield the same solution. To prevent this redundancy, a \emph{transition dominance} rule is applied, ensuring that any specific solution is generated by only a single sequence of tasks.
Formally, we define $\eta(S)$ as the subset of available operations that are legally permitted to expand $T_S$. An operation only qualifies for $\eta(S)$ if it strictly extends the makespan or resolves ties using $\Lambda$:

\begin{definition}
\label{def:transitiondominance}
An operation $o \in \varepsilon(S)$ belongs to $\eta(S)$ if it satisfies one of the following conditions:
\begin{itemize}
    \item $\psi_o + p_o > C_{\max}(S)$
    \item $\psi_o + p_o = C_{\max}(S) \land m(o) > \Lambda(S)$
\end{itemize}
\end{definition}

\begin{example}
\cref{tab:nextop} and \cref{tab:samemachine} illustrate the two cases of \cref{def:transitiondominance} on a state $S$.
The darker color is used for scheduled operations, while the hatched light color is for unscheduled ones.
In both cases, the ID of the last machine is 2.

In \cref{tab:nextop}, the set of available operations is
$\varepsilon(S)=\{o_1,\, o_5,\, o_8\}$.
Operations $o_1$ and $o_8$ satisfy the first condition.
Operation $o_5$ does not satisfy any conditions because
$\psi_{o_5}+p_{o_5}=2+1=3 < C_{\max}(S)=4$.
Therefore it is excluded from $\eta(S)$ and we obtain
$\eta(S)=\{o_1,\, o_8\}$.

In \cref{tab:samemachine}, assuming another set of operations, the available operations are
$\varepsilon(S)=\{o_3,\, o_6,\, o_8\}$.
Operations $o_3$ and $o_6$ satisfy the first condition.
Operation $o_8$ satisfies the second condition of \cref{def:transitiondominance} since
$m(o_8)=3 > \Lambda(S)=2$.
Consequently, all available operations satisfy the dominance rule and
$\eta(S)=\varepsilon(S)=\{o_3,\, o_6,\, o_8\}$.
\end{example}

The \emph{transition function}, denoted $\mathit{Transition}$, is defined by the operations in the set $\eta(S)$.
Given a state $S$ and an operation $o \in \eta(S)$, the transition function returns a new state $S' = \langle \psi', \Lambda', \mathit{done}' \rangle$.
First, the set of scheduled operations is updated: $\mathit{done}' = \mathit{done}(S) \cup \{o\}$.
Second, the earliest starting times $\psi$ are updated to $\psi'$ using the $\mathit{UpdateEst}$ function. This function performs the following updates:
\begin{enumerate}
    \item It updates the earliest start time of all unscheduled operations on the same machine as $o$:
    \begin{equation}
        \psi'_{o'} = \max(\psi_{o'}, \psi_o + p_o) \quad \forall o' \notin \mathit{done}' \mid m(o') = m(o) \label{eq:updateEst1}
    \end{equation}
    \item For each operation $o'$ whose $\psi'_{o'}$ has been increased, it recursively updates its successors in the precedence graph $G$:
    \begin{equation}
        \psi'_{o''} = \max(\psi_{o''}, \psi'_{o'} + p_{o'}) \quad \forall o'' \in \mathit{succs}(o') \label{eq:updateEst2}
    \end{equation}
\end{enumerate}
Third, the ID of the last machine is updated: $\Lambda' = m(o)$.
The transition function thus returns the new state $S' = \mathit{Transition}(S, o) = (\psi', \Lambda', \mathit{done}')$.
From a state $S$ and an operation $o\in \eta(S)$, the \emph{transition cost function} denoted $h$, associated with the transition is the makespan of the new state minus the makespan of the current state, i.e., $h(S,o)=C_{\max}(S')-C_{\max}(S)$. 
The \emph{domain function} of a state $S$, denoted $d(S)$, is the set of all operations that can directly extend the schedule.
The size of this set represents the branching factor of our search tree.
In the DP formulation of the JSP in \cite{GromichoHST12,HoornNOG17}, it is set $d(S)=\eta(S)$ for all states $S$.

\begin{figure}[t]
  \centering
  \begin{minipage}[b]{0.45\textwidth}
    \centering
    \begin{tikzpicture}

    \coordinate (origin) at (0,0);
    
    \begin{scope}[shift={(origin)}]
      % ======= BACKDROP: time grid & axes =======
      \foreach \x in {0,...,11} {
        \draw[very thin,gray!40] (-1.8+\x*0.5,0.2) -- (-1.8+\x*0.5,-1.5*\lanesep-0.9);
      }
      \foreach \x in {0,...,11} {
        \node[font=\scriptsize] at (-1.8+\x*0.5,-2.7) {\x};
      }
      \node[anchor=west] at (-2.6,0) {\large M1};
      \node[anchor=west] at (-2.6,-\lanesep) {\large M2};
      \node[anchor=west] at (-2.6,-2*\lanesep) {\large M3};

      \draw[thick] (-1.8,0.2) rectangle (3.7,-1.5*\lanesep-0.9);

      % Operations
      % Operations
      \Op{1}{0}{2}{2}{1}
      \Opv{1}{2}{3}{1}{1}
      % \Opv{1}{7}{1}{3}{3}

      \Op{2}{0}{4}{3}{1}
      % \Opv{2}{5}{2}{1}{2}
      % \Opv{2}{3}{4}{2}{3}

      \Opv{3}{2}{1}{2}{2}
      \Opv{3}{4}{3}{3}{2}
      % \Opv{3}{7}{2}{1}{3}

    \end{scope}

      % Legend
      \begin{scope}[shift={(origin)}]
        \draw[opJ1] (-2.5,-3.5) rectangle +(1,0.5);\node[anchor=west] at (-1.5,-3.25) {Job 1};
        \draw[opJ2] (-0.3,-3.5) rectangle +(1,0.5);\node[anchor=west] at (0.7,-3.25) {Job 2};
        \draw[opJ3] (1.8,-3.5) rectangle +(1,0.5);\node[anchor=west] at (2.8,-3.25) {Job 3};
      \end{scope}
      
    \end{tikzpicture}
    \caption{Illustration of $\varepsilon(S)$ and $\eta(S)$}
    \label{tab:nextop}
  \end{minipage}
  \hfill
  \begin{minipage}[b]{0.45\textwidth}
    \centering
    \begin{tikzpicture}
    \coordinate (origin) at (0,0);
    
    \begin{scope}[shift={(origin)}]
      % ======= BACKDROP: time grid & axes =======
      \foreach \x in {0,...,11} {
        \draw[very thin,gray!40] (-1.8+\x*0.5,0.2) -- (-1.8+\x*0.5,-1.5*\lanesep-0.9);
      }
      \foreach \x in {0,...,11} {
        \node[font=\scriptsize] at (-1.8+\x*0.5,-2.7) {\x};
      }
      \node[anchor=west] at (-2.6,0) {\large M1};
      \node[anchor=west] at (-2.6,-\lanesep) {\large M2};
      \node[anchor=west] at (-2.6,-2*\lanesep) {\large M3};

      \draw[thick] (-1.8,0.2) rectangle (3.7,-1.5*\lanesep-0.9);

      % Operations
      % Operations
      \Op{1}{0}{2}{2}{1}
      \Op{1}{2}{3}{1}{1}
      %\Op{1}{5}{2}{3}{3}

      \Op{2}{0}{4}{3}{1}
      \Op{2}{5}{2}{1}{2}
      \Opv{2}{7}{4}{2}{3}

      \Op{3}{2}{1}{2}{2}
      \Opv{3}{4}{3}{3}{2}
      \Opv{3}{7}{2}{1}{3}

    \end{scope}

      % Legend
      \begin{scope}[shift={(origin)}]
        \draw[opJ1] (-2.5,-3.5) rectangle +(1,0.5);\node[anchor=west] at (-1.5,-3.25) {Job 1};
        \draw[opJ2] (-0.3,-3.5) rectangle +(1,0.5);\node[anchor=west] at (0.7,-3.25) {Job 2};
        \draw[opJ3] (1.8,-3.5) rectangle +(1,0.5);\node[anchor=west] at (2.8,-3.25) {Job 3};
      \end{scope}

    \end{tikzpicture}
    \caption{Illustration of $\varepsilon(S)$ and $\eta(S)$}
    \label{tab:samemachine}
  \end{minipage}
\end{figure}

\subsection{State Dominance} \label{subsec:dominance}

Another dominance rule used in \cite{GromichoHST12} is based on the notion of the earliest completion time of the operation $o$ following the state $S$ denoted $\alpha_S(o)$ and specified by: 
 \begin{equation}
\label{eq:est}
    \alpha_S(o) = 
    \begin{cases}
    & \psi_o + p_o \quad \text{if } o \in \eta(S) \\
    & C_{\max}(S) + p_o \quad \text{ otherwise}
    \end{cases}
\end{equation}

\begin{definition}\cite{GromichoHST12}
\label{def:state_dom}
    A state $S_1$  dominates a state $S_2$, denoted $S_1 \preceq S_2$ if 
\begin{equation}
    \begin{cases}
    & \mathit{done}(S_1) = \mathit{done}(S_2) \\
    & \alpha_{S_1}(o) \leq \alpha_{S_2}(o) \quad \forall o \in \varepsilon(S_1) = \varepsilon(S_2)
    \end{cases}
    \end{equation}
\end{definition}

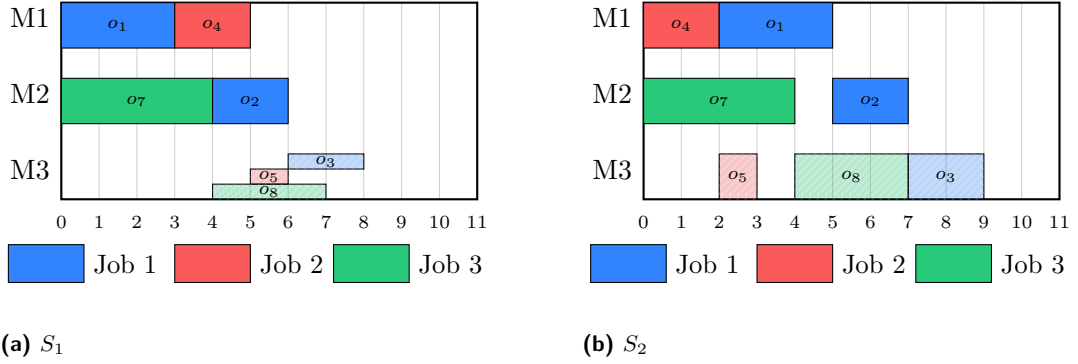
\begin{figure}[!ht]
  \centering
  \begin{subfigure}[b]{0.45\textwidth}
    \centering
    \begin{tikzpicture}

    \coordinate (origin) at (0,0);
    
    \begin{scope}[shift={(origin)}]
      % ======= BACKDROP: time grid & axes =======
      \foreach \x in {0,...,11} {
        \draw[very thin,gray!40] (-1.8+\x*0.5,0.2) -- (-1.8+\x*0.5,-1.5*\lanesep-0.9);
      }
      \foreach \x in {0,...,11} {
        \node[font=\scriptsize] at (-1.8+\x*0.5,-2.7) {\x};
      }
      \node[anchor=west] at (-2.6,0) {\large M1};
      \node[anchor=west] at (-2.6,-\lanesep) {\large M2};
      \node[anchor=west] at (-2.6,-2*\lanesep) {\large M3};

      \draw[thick] (-1.8,0.2) rectangle (3.7,-1.5*\lanesep-0.9);

      % Operations
      \Op{1}{0}{3}{1}{1}
      \Op{1}{3}{2}{2}{1}
      % \Opv{1}{7}{1}{3}{3}

      \Op{2}{0}{4}{3}{1}
      \Op{2}{4}{2}{1}{2}
      % \Opv{2}{3}{4}{2}{3}

      \Ops{3}{5}{1}{2}{2}{1}{3}
      \Ops{3}{4}{3}{3}{2}{0}{3}
      \Ops{3}{6}{2}{1}{3}{2}{3}

    \end{scope}

      % Legend
      \begin{scope}[shift={(origin)}]
        \draw[opJ1] (-2.5,-3.5) rectangle +(1,0.5);\node[anchor=west] at (-1.5,-3.25) {Job 1};
        \draw[opJ2] (-0.3,-3.5) rectangle +(1,0.5);\node[anchor=west] at (0.7,-3.25) {Job 2};
        \draw[opJ3] (1.8,-3.5) rectangle +(1,0.5);\node[anchor=west] at (2.8,-3.25) {Job 3};
      \end{scope}
      
    \end{tikzpicture}
    \caption{$S_1$}
    \label{tab:statedomine}
  \end{subfigure}
  \hfill
  \begin{subfigure}[b]{0.45\textwidth}
    \centering
    \begin{tikzpicture}

\coordinate (origin) at (0,0);
    
    \begin{scope}[shift={(origin)}]
      % ======= BACKDROP: time grid & axes =======
      \foreach \x in {0,...,11} {
        \draw[very thin,gray!40] (-1.8+\x*0.5,0.2) -- (-1.8+\x*0.5,-1.5*\lanesep-0.9);
      }
      \foreach \x in {0,...,11} {
        \node[font=\scriptsize] at (-1.8+\x*0.5,-2.7) {\x};
      }
      \node[anchor=west] at (-2.6,0) {\large M1};
      \node[anchor=west] at (-2.6,-\lanesep) {\large M2};
      \node[anchor=west] at (-2.6,-2*\lanesep) {\large M3};

      \draw[thick] (-1.8,0.2) rectangle (3.7,-1.5*\lanesep-0.9);

      % Operations
      \Op{1}{0}{2}{2}{1}
      \Op{1}{2}{3}{1}{1}
      % \Opv{1}{7}{1}{3}{3}

      \Op{2}{0}{4}{3}{1}
      \Op{2}{5}{2}{1}{2}
      % \Opv{2}{3}{4}{2}{3}

      \Opv{3}{2}{1}{2}{2}
      \Opv{3}{4}{3}{3}{2}
      \Opv{3}{7}{2}{1}{3}

    \end{scope}

      % Legend
      \begin{scope}[shift={(origin)}]
        \draw[opJ1] (-2.5,-3.5) rectangle +(1,0.5);\node[anchor=west] at (-1.5,-3.25) {Job 1};
        \draw[opJ2] (-0.3,-3.5) rectangle +(1,0.5);\node[anchor=west] at (0.7,-3.25) {Job 2};
        \draw[opJ3] (1.8,-3.5) rectangle +(1,0.5);\node[anchor=west] at (2.8,-3.25) {Job 3};
      \end{scope}
      
    \end{tikzpicture}
    \caption{$S_2$}
    \label{tab:statedominated}
  \end{subfigure}
  \caption{Example of state dominance: $S_1 \preceq S_2$}
  \label{tab:dominance}

\end{figure}

\begin{example}
\cref{tab:dominance} illustrates \cref{def:state_dom}, that is $S_1 \preceq S_2$
with $S_1$ depicted in \cref{tab:statedomine} and $S_2$ in \cref{tab:statedominated}.
For $S_2$, $o_{5} \notin \eta(S_2)$, and therefore $\alpha_{S_2}(o_{5}) = 8$ from \cref{eq:est}.
It follows that $\alpha_{S_1}(o_{3}) = 8 < 9=\alpha_{S_2}(o_{3})$, $\alpha_{S_1}(o_{5}) = 6 < 8=\alpha_{S_2}(o_{5})$ and $\alpha_{S_1}(o_{8}) = 7 =\alpha_{S_2}(o_{8})$. Thus $S_1 \preceq S_2$.
\end{example} 

Another state dominance is proposed in \cite{HoornNOG17}.
A state $S$ is dominated when there is at least one machine with an unscheduled operation belonging to $\eta(S)$ and all operations of this machine in $\eta(S)$ cannot start before $C_{\max}(S)$.
Such states can be discarded, since the available operation must be scheduled at its earliest start time. 
Formally a state $S$ is dominated if there is a machine $M_j$ and an operation $o \notin \eta(S)$ with $ m(o) = M_j$ such that
\begin{equation}
    \alpha_S(o') = C_{\max}(S) + p_{o'} \quad \forall o' \in \varepsilon(S) \quad \land \quad m(o') = M_j.
    \label{eq:reduceSpace}
\end{equation}

\section{Anytime Column Search (ACS)}\label{sec:acs}

The DP model starts from the root state $\hat{r}=\langle(0,\dots,0), -1, \emptyset\rangle$.
The solution space is then explored layer-wise from the root, as in decision diagrams \cite{BergmanCHH16}, to apply and detect all state dominance rules.
A first drawback of this approach is that the optimal solution (i.e., the shortest path in the transition graph) is discovered only at the very end, which leads to poor anytime behavior.
A second drawback is that a branch-and-bound search would enable better filtering of the states by pruning nodes that are provably unable to improve the incumbent solution.
To address these issues, we propose exploring the search space using the Anytime Column Search (ACS) \cite{vadlamudi2012anytime}, as depicted in \cref{alg:acs}, rather than a layer-wise exploration.

In ACS, to limit the size of the search space, only a fixed number of nodes, denoted $W$, are expanded at each layer, similarly to a beam search.
In \cite{vanHoorn}, the lower bound of a state is computed with the Jackson Preemptive Schedule (JPS) \cite{jackson1955scheduling}. The JPS works as follows.
On a machine, at least one available operation with the minimum latest completion time is scheduled. This operation is processed either up to its completion or until a more urgent operation becomes available. The function $\mathit{LowerBoundJPS}(S)$ denotes the lower bound on the current state $S$ using JPS and taking the maximum JPS makespan across all machines. The priority used to select the nodes to expand in ACS is defined by $f(S) = \mathit{LowerBoundJPS}(S)$.

\cref{alg:acs} receives as input an instance $I$ of JSP or OSP and the parameter $W$, and provides the optimal solution within the time limit or an approximate solution otherwise.
An array of empty queues of states (one queue per decision layer) sorted in increasing value of $f$ is initialized at Line \ref{initQueue}.
The first queue links to the decisions on the root state, is filled in the loop of Line \ref{firstDecision} with the transition from the root and their domain $d(\hat{r})$.
In the loop of Line \ref{loopDecision}, the best $W$ non-dominated states of each layer are made in the loop of Line \ref{loopVariables}.
%Indeed, the loop of Line \ref{selectCandidates} selects $W$ non-dominated and promising states for each layer.
The function $\mathit{isNotDominated}$ at Line \ref{line:dominance} checks if the state is not dominated by any other state of the same layer, i.e., $\forall S' \in X[l] \mid S' \npreceq S$, as defined in \cref{def:state_dom} or if it is not dominated according to the \cref{eq:reduceSpace}.
The loop at Line \ref{fileNext} fills the queue associated with the next layer with the transition states of the selected states of the previous decision layer and their domains. During the transition, the lower bound for the state will be computed. 
The loop of Line \ref{checkSol} checks if there exists an improving solution in the queue of the last decision. 

\paragraph{Implementation of state dominance.}
Following \cite{coppe_dominance}, state dominance is implemented via a dedicated map indexed by the \emph{dominance key} of a state, defined as its set of scheduled operations $\mathit{done}(S)$.
For each dominance key, the map stores all non-dominated states seen so far with that exact set of scheduled operations.
When a state $S$ is dequeued and checked by $\mathit{isNotDominated}$, it is compared against the map entry for its dominance key; if any stored state $S'$ satisfies $S' \preceq S$, the state is discarded.
Crucially, states remain in the map even after being dequeued and expanded, so a state created later that is dominated by an already-expanded node is still correctly discarded.
However, the converse is not guaranteed: a state that has already been expanded may, in a later iteration, be dominated by a newly created state.
This situation can arise because ACS does not develop states strictly layer-by-layer as the original DP approach does \cite{GromichoHST12,HoornNOG17}; consequently, some nodes may be explored that a purely layer-wise traversal would have pruned.
This is an inherent tradeoff of ACS: accepting occasional redundant expansions in exchange for the ability to dive heuristically toward high-quality solutions early, which in turn enables tighter bound pruning across subsequent iterations.

\begin{algorithm2e}[t]
\caption{\textbf{AnytimeColumnSearch$(I, W)$}}\label{alg:acs}
\KwIn{An instance $I$ with $N$ operations and a maximal width $W$}
\KwOut{The optimal sequence or an approximation solution}
%$X \gets [\,]$  \tcp*{Each level is sorted according to $f$}
\For{$l = 1$ \KwTo $N$}{\label{initQueue}
    $X[l] \gets \emptyset$ \tcp*{empty queue of states sorted according to $f$}
}
$\mathit{ub} \gets \mathit{horizon}$ \;
$\hat{r} \gets \langle(0,\dots,0), -1, \emptyset\rangle$\;

\ForEach{$o \in d(\hat{r})$}{\label{firstDecision}
    $X[1] \gets X[1] \cup \{t(\hat{r}, o)\}$ \;
}
\While{$X \neq \emptyset$}{\label{loopDecision}
    \For{$l = 1$ \KwTo $N -1$}{\label{loopVariables}
        $\mathit{candidates} \gets \emptyset$\;
        \While{$|\mathit{candidates}| < W \wedge X[l]\neq \emptyset$}{\label{selectCandidates}
            $S \gets \mathit{dequeueFirst}(X[l])$\;
            \If{$\mathit{isNotDominated}(S) \land  \mathit{LowerBoundJPS}(S) < \mathit{ub}$}{\label{line:dominance}
                $\mathit{candidates} \gets \mathit{candidates} \cup \{S\}$ \;
            }
            
        }
        \For{$S \in \mathit{candidates}$}{
            \ForEach{$o \in d(S)$}{\label{fileNext}
                $S' \gets \mathit{Transition}(S, o)$ \; \label{line:ACSTransition}
                $X[l+1] \gets X[l+1] \cup \{S'\}$\;
                
            }
        }
    }
    
    \While{$X[N] \neq \emptyset$}{\label{checkSol}
        $S \gets \mathit{dequeueFirst}(X[N])$ \;
        \If{$C_{\max}(S) < \mathit{ub}$}{
            $\mathit{ub} \gets C_{\max}(S)$ \;
        }
                
    }
}
\end{algorithm2e}

\section{Hybrid DP-CP Model for the PSSP}\label{sec:hybrid}

In the PhD thesis of A. van Hoorn \cite{vanHoorn}, the idea of using a filtering algorithm called the \emph{parallel head–tail adjustment} \cite{brinkkotter2001solving} is briefly mentioned. In the constraint programming community, this algorithm is more commonly referred to as a type of \emph{edge-finding algorithm} \cite{vilim2007global,Martin1996ANA,baptiste2001constraint}.
Such algorithms allow the detection of precedence constraints. Van Hoorn informally explains that these precedences could be used to prune transitions further by allowing only expansions for which all predecessors are already in the partial solution \cite{vanHoorn}.

Our contribution with respect to this idea is twofold:

\begin{itemize}
    \item Use of full CP propagation. Instead of relying on the parallel head–tail adjustment, we invoke a CP solver implementing all the filtering algorithms from \cite{vilim2007global} for the \textsc{NoOverlap} constraint on each machine, including \emph{overload checking}, \emph{edge-finding}, \emph{detectable precedences}, and \emph{not-first/not-last} rules. Petr Vilím proved \cite{vilim2007global} that after reaching a fixpoint, all detectable precedences can be retrieved from the propagated time windows by verifying whether the earliest completion time of one activity exceeds the latest start time of another (denoted $\lessdot$ in the following). By computing tighter time windows through CP propagation and identifying additional precedence relations, the explored search space can therefore be further reduced.
  \item  Formalization of the approach. We fully formalize this idea and provide a precise transition function based on calls to the CP solver, thereby turning the previously informal suggestion into a well-defined algorithmic framework applicable to any scheduling problem with general precedences.
\end{itemize}

\subsection{Extended State and Update of the Domain Function}

The state definition from \cite{GromichoHST12} is extended to explicitly represent the set of precedences (both original from the problem and those dynamically discovered through CP time-window filtering).
We denote by $S=\langle \psi, \Lambda, \mathit{done}, \delta \rangle$ the extended state definition, where $\delta(S)$ denotes the set of precedence constraints currently known.
Given an operation $o\in O$, we denote by $\delta(S,o)$ the set of predecessors of $o$ in the precedence relations of $\delta(S)$, i.e.,
$\delta(S,o)=\{\,o'\in O \mid (o',o)\in \delta(S)\,\}$. At the root state, the set $\delta$ is initialized to $E$, the initial precedence graph of the problem. 

The domain function $d(S)$ is adapted to take $\delta(S)$ into account.  
Only operations whose predecessors are all already scheduled in $\delta(S)$ can be appended, i.e.,  
$\{o \in \eta(S) \mid \delta(S,o) \subseteq \mathit{done}(S)\}$.  
This restriction reduces the size of the domain function and, consequently, the branching factor and the size of the search space explored by the DP approach.

\subsection{Transition with Constraint programming (CP)}

The transition function relying on CP is given in \cref{alg:transition} and replaces the transition function in ACS at line \ref{line:ACSTransition}.  
Given an extended state $S$, an operation $o \in \eta(S)$, and an upper bound on the makespan, it returns a new state $S'$ in which the set of precedences includes both the precedence induced by the decision to append operation $o$ to the sequence and the detectable precedences identified after CP filtering.

Internally, this function makes use of a CP model for our scheduling problem given in \cref{alg:cp}, further restricted by the current state. This function returns a constraint satisfaction problem (CSP) as a pair $P = \langle \mathcal{I}, \mathcal{C} \rangle$ where $\mathcal{I}$ is a mapping from operations to their corresponding CP interval variables \cite{LaborieR08,abs-2508-01751} and $\mathcal{C}$ the set of CP constraints. CP interval variable ranges are initialized by the function 
$\mathit{GenerateIntervals}$ as follows.

\begin{itemize}
\item For operations in $\mathit{done}(S)$, the start time is fixed to their starting time in the state.
\item For operations in $\eta(S)$, the minimum start time is set to their earliest starting time in $S$.
\item For operations in $\varepsilon(S)\setminus \eta(S)$, the minimum start time is set to $C_{\max}(S)$.
\item For operations in $\mathit{done}(S)$, the end time is fixed to their starting time plus their duration. 
\item For operation in $O\setminus (\mathit{done}(S))$, the maximum end time is fixed to the $\mathit{ub}$.
\end{itemize}

The precedence constraints $\delta(S)$ and disjunction constraints for any machine and/or job are collected into a set $\mathcal{C}$ (Lines \ref{disjunctions}, \ref{disjunctions_job}).

\begin{algorithm2e}[!ht]
\caption{\textbf{InitializeCSP$(S, \mathit{ub})$}}\label{alg:cp}
\KwIn{A state $S$, an upper bound $\mathit{ub}$.}
\KwOut{A CSP $P = \langle \mathcal{I}, \mathcal{C} \rangle$ where $\mathcal{I}$ maps operations to interval variables and $\mathcal{C}$ is a set of constraints.}

$\mathcal{I} \gets  \mathit{GenerateIntervals}(S, \mathit{ub})$ \;
$\mathcal{C} \gets \emptyset$ \;

\For {$(o',o'') \in \delta(S)$ } {\label{precedences}
    $\mathcal{C} \gets \mathcal{C} \cup \{\mathit{StartBeforeEnd}(\mathcal{I}(o'), \mathcal{I}(o''))\}$\;
}
\For{$j = 1$ \KwTo $m$} { \label{disjunctions}
    $\mathcal{C} \gets \mathcal{C} \cup \{\mathit{NoOverlap}(\{\mathcal{I}(o') \mid  o' \in O \land m(o')=j \})\}$\;
}
\For{$i = 1$ \KwTo $n$} { \label{disjunctions_job}
    $\mathcal{C} \gets \mathcal{C} \cup \{\mathit{NoOverlap}(\{\mathcal{I}(o') \mid  o' \in O \land j(o')=i \})\}$\;
}

\Return $\langle \mathcal{I}, \mathcal{C} \rangle$ \;

\end{algorithm2e}

\begin{algorithm2e}[!hb]
\caption{\textbf{TransitionCP$(S, o, \mathit{ub})$}}\label{alg:transition}
\KwIn{A state $S$, an operation $o$ and an upper bound $\mathit{ub}$.}
\KwOut{The new state $S'$.}
$\psi' \gets \mathit{UpdateEst}(\psi, o), \Lambda' = m(o), \mathit{done}' = \mathit{done}(S)\cup\{o\}, \delta' \gets \delta(S)$\;
$S' \gets \langle \psi', \Lambda', \mathit{done}', \delta' \rangle $\;
$P = \langle \mathcal{I}, \mathcal{C} \rangle \leftarrow \mathit{InitializeCSP}(S', \mathit{ub})$ \;\label{cspinit}
\If{$\mathit{Fixpoint}(P)=\mathit{Failure}$}{ \label{fixp}
            \Return $\mathit{None}$.
}
$\delta' \gets \delta(S)$\;
\ForEach{$o' \in O$}{\label{precds}
    \ForEach{$o'' \in O \setminus \{o'\}$}{
        \If{$\mathcal{I}(o'') \lessdot \mathcal{I}(o')$}{
            $\delta' \gets \delta' \cup \{(o'',o')\}$\;
        }
    }
}
\Return $S'=\langle \psi',\Lambda',\mathit{done}', \delta'\rangle$\;
\end{algorithm2e}

In \cref{alg:transition}, after the initialization of the CSP at Line \ref{cspinit}, the fixed point is computed at Line \ref{fixp}.
The loop of Line \ref{precds} analyzes the interval at the fixed point to retrieve the detectable precedences to be added in $\delta'$.
The value of $\psi'$ is updated with the function $\mathit{UpdateEst}$ described in \cref{subsec:trans_dominance} with the \cref{eq:updateEst1} and the \cref{eq:updateEst2}. 

Notice that we do not update $\psi$ with the earliest starting time of the interval variables in $\mathcal{I}$ because we want to preserve the dominance rules defined in \cref{subsec:dominance}. If we were to update it, the computation of $\eta$ would become invalid.

\subsection{Lower Bound With Constraint Programming}

The lower bound is used in state-space exploration to reduce the search space, thereby speeding up and tightening the problem bounds.
We adopt a lower bound similar to that in \cite{carlier1989algorithm}, which uses a dichotomic approach with filtering algorithms, denoted $\mathit{LowerBoundCP}$. This function replaces the function $\mathit{LowerBoundJPS}$ at line \ref{line:dominance} in the \ref{alg:acs}.
\cref{alg:lb} receives as input a state $S$ and an upper bound $\mathit{ub}$ and computes a lower bound from $S$ to be used in the anytime column search.
The initial lower bound is computed with the Jackson preemptive scheduling procedure at Line \ref{jackson}, and the function $\mathit{InitializeCSP}(S, \mathit{mid})$ returns the pair $\langle \mathcal{I}, \mathcal{C} \rangle$ as described in \cref{alg:cp} with $\mathit{mid}$ as an upper bound.
The function $\mathit{Fixpoint}$ computes the fix-point and returns $\mathit{Failure}$ if an inconsistency is detected.
The lower bound plays a central structural role in anytime column search (ACS) since it drives convergence, pruning, and the anytime behavior.

\begin{algorithm2e}[!ht]
\caption{\textbf{LowerBoundCP$(S, \mathit{ub})$}}\label{alg:lb}
\KwIn{A state $S$, an upper bound $\mathit{ub}$.}
\KwOut{The lower bound from the state $S$.}

$\mathit{lb} \gets \mathit{JacksonPreemptive}(S, \mathit{ub})$\; \label{jackson}

\While{$\mathit{lb} < \mathit{ub}$}{
    $\mathit{mid} \gets \lfloor (\mathit{lb} + \mathit{ub})/2 \rfloor$\;
    $P = \langle \mathcal{I}, \mathcal{C} \rangle \gets \mathit{InitializeCSP}(S, \mathit{mid}) $\;
    \If{$\mathit{Fixpoint}(P) = \mathit{Failure}$}{
        $\mathit{lb} \gets \mathit{mid} + 1$\;
    }
    \Else{
        $\mathit{ub} \gets \mathit{mid}$\;
    }
}

\Return $\mathit{lb}$\;
\end{algorithm2e}

\section{Large Neighborhood Search (LNS)}\label{sec:lns}

Given a solution to a problem, the generic form of the large neighborhood search (LNS) builds a large neighborhood of this solution and uses a solver to explore the neighborhood and improve the solution \cite{shaw1998using}.
To build a neighborhood of a solution, a fraction of the incumbent solution is relaxed, and the corresponding problem is passed to the solver. 
From the acyclic graph associated with the incumbent solution, a fraction of the precedence relations obtained during the search is relaxed, and the subproblem corresponding to this state is launched with an upper bound to allow the search to improve the solution.
To explore the search space of this subproblem, we perform an A* search \cite{hart1968formal} rather than an ACS. 
A* is a best-first search (BFS) algorithm that finds the optimal solution when finished (without improving it over time like ACS).
\cref{alg:lns} receives as input an instance $I$ of the problem, an initial solution $S_{best}$, and returns an improved solution.
The CSP $P = \langle \mathcal{I}, \mathcal{C} \rangle$ is initialized with the root state $\hat{r}$ and the upper bound $C_{\max}(S_{best})-1$ using the function of \cref{alg:cp} at Line \ref{initialcsp}.
If the initialization fails (i.e., the fixpoint detects a failure), the flag $\mathit{isOptimal}$ is set to true, meaning no better solution exists, and the function returns the current solution.
A subset of the precedences from the current solution, denoted $\delta_{kept} \subset \delta(S_{best}) \setminus E$, is selected to be maintained (relaxed precedences are ignored) in the function $\mathit{SelectSubsetPrecedences}$. These are added to the set of constraints $\mathcal{C}$ in the loop of Line \ref{fixedPrecedences}.
The CSP $P$ is then used to define the search space for an A* search, $\mathit{AStarSearch}(P)$, which explores the neighborhood.
If a solution $S_{new}$ is found, it replaces $S_{best}$.

\begin{algorithm2e}[!ht]
\caption{\textbf{LNS$(I, S_{best})$}}\label{alg:lns}
\KwIn{An instance $I$ and an initial solution $S_{best}$.}
\KwOut{An improved solution $S_{best}$ and a status $\mathit{isOptimal}$.}
$\mathit{isOptimal} \gets \mathit{false}$\;
\While{stopping criterion not met}{
    $P = \langle \mathcal{I}, \mathcal{C} \rangle \gets \mathit{InitializeCSP}(\hat{r}, C_{\max}(S_{best})-1) $\;\label{initialcsp}
    \If{$\mathit{Fixpoint}(P) = \mathit{Failure}$}{
        $\mathit{isOptimal} \gets \mathit{true}$\;
        \textbf{break}\;
    }
    $\delta_{kept} \gets \mathit{SelectSubsetPrecedences}(S_{best})$\; 
    \ForEach{$(o, o') \in \delta_{kept}$}{\label{fixedPrecedences}
        $\mathcal{C} \gets \mathcal{C} \cup \{\mathit{StartBeforeEnd}(\mathcal{I}(o), \mathcal{I}(o'))\}$\;
    }
    
    $S_{new} \gets \mathit{AStarSearch}(P)$\; 
    \If{$S_{new} \neq \mathit{null}$}{
         $S_{best} \gets S_{new}$\;
    }
}
\Return $S_{best}, \mathit{isOptimal}$ \;
\end{algorithm2e}
After a large number of restarts of the A* search without improving the solution, the fraction of precedence relaxed from the current solution is reduced. 
If this percentage falls below a threshold, it returns to its initial value. 
This feature is known as adaptive LNS.  

\section{Experimental Results}\label{sec:xps}

In this section, we report an empirical evaluation of different configurations of our DP-CP model on six well-known benchmarks, split as follows: three for JSP and three for OSP. 
All instances are available \footnote{\url{https://github.com/ScheduleOpt/benchmarks/}}. 
For the JSP, we use the benchmark suites of Fisher and Thompson \cite{fisher1963job}, Lawrence \cite{lawrance1984resource}, and  Applegate and Cook \cite{applagate1991computational}.
The benchmarks used for the OPS are from Taillard \cite{taillard1993benchmarks},  Gu\'eret and Prins \cite{gueret1999new}, and Brucker, Hurink, Jurisch, and Wöstmann \cite{brucker1997branch}.
The implementation of the DP part of the approach uses an open-source Java version~\cite{Schaus2026DDOLib} of \cite{GillardSC20} and the CP part uses MaxiCP \cite{maxicp}, an open source CP solver\footnote{\url{http://www.maxicp.org}}.
The different search strategies available in DDOLib were tested in our DP model, among which the decision diagram–based branch and bound (DD-B\&B) search \cite{GillardSC20}, the A* search, and the anytime column search (ACS) \cite{vadlamudi2012anytime}.
We have not found a suitable merge operator for the DD-B\&B to speed up the approach, and the quality of the initial upper bound affects its performance, as in A*.
The improvement over time of ACS provides a good initial solution quickly, even with a large initial upper bound.
%In the remaining experiments, we adopt ACS search across all DP models.
We set the parameter $W$ to 5 after preliminary experiments, which is a good compromise between the number of explored nodes and execution time.

The computational experiments reported in this section were run on an Intel(R) Xeon(R) Platinum 8160 CPU with 314GB of RAM, with a timeout of 3600 seconds.
The source code and instances are available \footnote{\url{https://anonymous.4open.science/r/DP-CP_JobShop-8C5E}}.
All acronyms used in the results are described in \cref{tab:model}.

\begin{table}[!ht]
    \centering
    \begin{tabular}{|p{0.2\textwidth}|p{0.7\textwidth}|}\hline
     Model    &  Description\\\hline
     DP-CP    &  Complete model \\
     DP-JPS   &  DP model without CP in transition and lower bound \\
     DP-CP-JPS & DP model with CP in transition but no CP in lower bound \\
     CP-S     & CP model with setTimes search strategy \\
     CP-R     & CP model with Rank search strategy \\
     LNS      &  LNS based on DP-CP with A* search, and the first solution provided by the complete model\\
     \hline
    \end{tabular}
    \caption{Models considered in the experiments}
    \label{tab:model}
\end{table}

We provide some answers to questions driven by our experiments.
These answers are based on detailed results presented afterward.
\begin{description}
\item[Question 1] Is an anytime column search necessary for the DP model and thus for the DP-CP hybridization model? \\
    \textbf{Answer:} Yes, since we notice a reduction in the number of explored nodes when compared to the state-of-the-art DP model, where a breadth-first search is used.
    
    \item[Question 2] Does hybridization DP-CP help to improve the individual approaches? \\
    \textbf{Answer:} Yes for both, since we notice a reduction in the number of explored nodes when compared to the individual approaches.
    
    \item[Question 3] Does the CP lower bound help the hybridization DP-CP to be competitive with the individual approaches? \\
    \textbf{Answer:} Yes, since a greater reduction of explored nodes is noticed with each ingredient.
    
    \item[Question 4] Does hybridization DP-CP help as a black box to explore the solution neighborhood? \\
    \textbf{Answer:} Yes, on both problems JSP and OSP.
\end{description}

\subsection{Question 1}
We compare the number of explored nodes of DP-JPS with the results reported from \cite{GromichoHST12}.
Despite the simplicity of the lower bound ($\mathit{LowerBoundJPS}$) coupled with ACS, the results reported in \cref{tab:SOTAvsACS} clearly show that the number of explored nodes is substantially reduced by the anytime column search used to find and prove the optimality. 
This is expected because the state-of-the-art approach uses a breadth-first search, and the JPS-based lower bound enables ACS to control the search and guarantee the quality of the current solution.        
\begin{table}[b]
    \centering
    \begin{tabular}{|c|r|r|}
         \hline
         Instances & State-of-the-art \cite{GromichoHST12}& DP-JPS \\
         \hline
         ft06& 30 409 & \textbf{427}\\
         la01& 63 170 946& \textbf{4 890} \\
         la02& 80 862 876 & \textbf{31 475}\\
         la03& 50 910 284 & \textbf{15 983}\\
         la04& 68 208 819& \textbf{28 491}\\
         la05& 40 229 147 & \textbf{2 180}\\
         \hline
    \end{tabular}
    \caption{Nodes explored: DP-JPS vs state-of-the-art \cite{GromichoHST12}}
    \label{tab:SOTAvsACS}
\end{table}

\subsection{Question 2}
\begin{figure}[t] 
    \begin{minipage}{0.49\textwidth}
        \centering
        \includegraphics{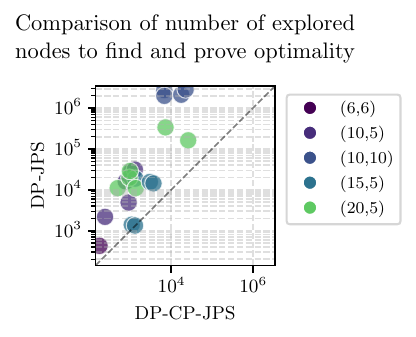}
        \caption{Nodes to optimality comparison. Labels $(X,Y)$ denote $X$ jobs $\times$ $Y$ machines. Dots: both solvers proved optimality; crosses: at least one solver timed out.}
        \label{fig:wocpVScp_nodes}
    \end{minipage}
    \hfill
    \begin{minipage}{0.49\textwidth}
        \centering        
        \includegraphics{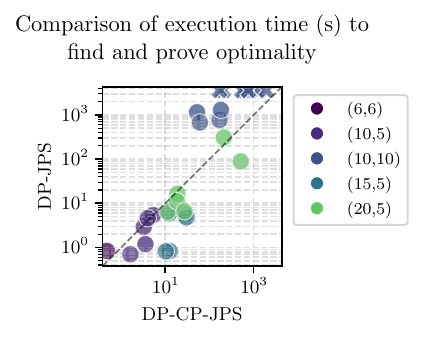}
        \caption{Time to optimality comparison. Labels $(X,Y)$ denote $X$ jobs $\times$ $Y$ machines. Dots: both solvers proved optimality; crosses: at least one solver timed out.}
        \label{fig:wocpVScp_time}
    \end{minipage}
\end{figure}

\paragraph*{DP-CP vs DP}
The DP model, denoted DP-JPS, is compared to our DP-CP-JPS model, referred to as DP-CP, without CP in the lower bound computation.
The DP-JSP model explores the search space with ACS but without CP in transition, and for the lower bounds computation ($\mathit{LowerBoundJPS}$ is used).
The models differ only in the transition phase, where the CP is used in DP-CP-JPS to reduce the branching factor of the DP formulation.  
\cref{fig:wocpVScp_nodes} compares the number of nodes explored needed to find and prove the optimality, while \cref{fig:wocpVScp_time} compares the time used to do so in JSP.
We observe a reduction in the number of nodes explored across all instances solved to optimality by the DP-CP-JPS model.
However, in some instances, the DP-CP-JPS model requires a slight increase in time.
The running time favors DP-JPS for instances solved in less than 100 seconds, and for hard instances, it generally favors the hybridization DP-CP-JPS (see \cref{fig:wocpVScp_time}).

\paragraph*{DP-CP vs CP}
We compare our hybridization DP-CP model against a pure CP model on JSP.
The CP model used the interval variables and the $\textsc{NoOverlap}$ global constraint available in MaxiCP \cite{maxicp}.
The strategy search used is the \textit{setTimes} \cite{le1995time} and \textit{Rank} (\cite{baptiste2001constraint} pages 130). 
\begin{figure}[b] 
    \begin{minipage}{0.49\textwidth}
        \centering \includegraphics{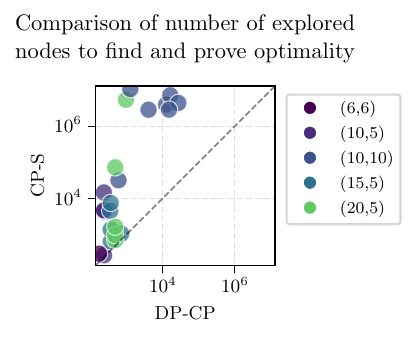}
        \caption{Nodes to optimality comparison (JSP, instances solved to optimality by both methods). Labels $(X,Y)$: $X$ jobs $\times$ $Y$ machines. Note: CP-S timed out on all $20{\times}10$ and $30{\times}10$ instances, which are therefore excluded. Dots: both proved optimality; crosses: at least one timed out.}
        \label{fig:nodes_setTimes}
    \end{minipage}
    \hfill
    \begin{minipage}{0.49\textwidth}
        \centering        \includegraphics{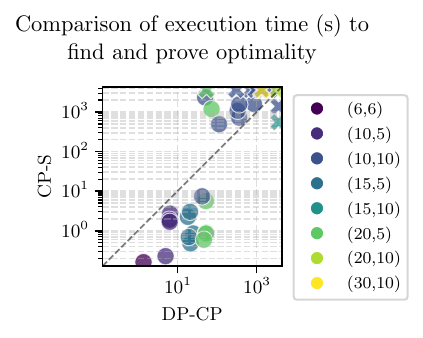}
        \caption{Time to optimality comparison (JSP, instances solved to optimality by both methods). Labels $(X,Y)$: $X$ jobs $\times$ $Y$ machines. Dots: both proved optimality; crosses: at least one timed out.}
        \label{fig:time_setTimes}
    \end{minipage}
\end{figure}
\cref{fig:nodes_setTimes} compares the number of explored nodes needed to find and prove the optimality of approaches, while \cref{fig:time_setTimes}  compares the execution time in seconds to find and prove the optimality on JSP instances where at least one finds and proves the optimality.
Undoubtedly, the hybridization DP-CP significantly reduces the number of explored nodes compared to the CP model, in order to find and prove optimality. 
This reduction in explored nodes reduced the running time of hard instances (instances that require more time to be solved to optimality) in favor of the hybridization DP-CP.
We also compare our model with a CP model using precedence branching. The results can be found in \cref{sec:results} of the appendix. It also compares the number of solved instances of our model with the two CP models in terms of time.

\begin{figure}[t] 
    \begin{minipage}{0.49\textwidth}
        \centering 
        \includegraphics{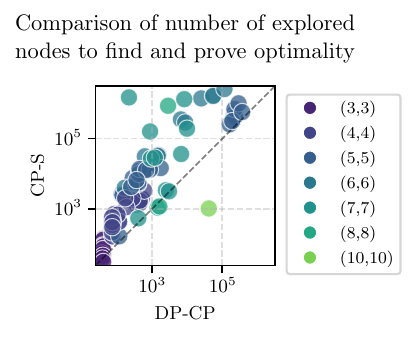}
        \caption{Nodes to optimality comparison (OSP). Labels $(X,Y)$: $X$ jobs $\times$ $Y$ machines. Dots: both proved optimality; crosses: at least one timed out.}
        \label{fig:os_nodes}
    \end{minipage}
    \hfill
    \begin{minipage}{0.49\textwidth}
        \centering        
        \includegraphics{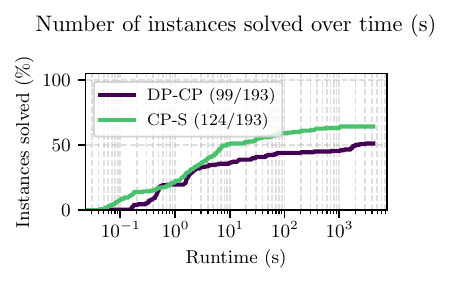}
        \caption{Solved instances over time (OSP)}
        \label{fig:os_time}
    \end{minipage}
\end{figure}

On OSP, we only apply the \textit{setTimes} search and compare our DP-CP hybridization model against CP-S.
\cref{fig:os_nodes} compares the number of explored nodes needed to find and prove the optimality, while \cref{fig:os_time} compares the number of common solved instances over time.
The hybridization reduces the number of nodes needed to find and prove the optimality of almost all OSP instances.
Despite the reduction, the CP-S model is still superior to the DP-CP model.
This is likely due to the hardness of the OSP, in which no order is specified for operations of the same job, as in JSP.

\subsection{Question 3}
\begin{figure}[!ht] 
    \begin{minipage}{0.49\textwidth}
        \centering \includegraphics{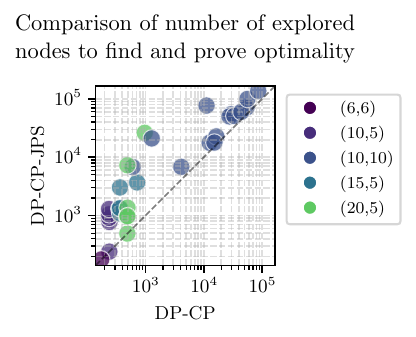}
        \caption{Nodes to optimality comparison. Labels $(X,Y)$: $X$ jobs $\times$ $Y$ machines. Dots: both proved optimality; crosses: at least one timed out.}
        \label{fig:dicho_nodes}
    \end{minipage}
    \hfill
    \begin{minipage}{0.49\textwidth}
        \centering        \includegraphics{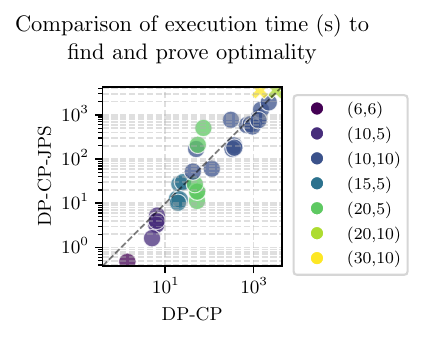}
        \caption{Time to optimality comparison. Labels $(X,Y)$: $X$ jobs $\times$ $Y$ machines. Dots: both proved optimality; crosses: at least one timed out.}
        \label{fig:dicho_times}
    \end{minipage}
\end{figure}

To appreciate the impact of the CP on the computation of the lower bound, we compare our model, DP-CP, with its variant, DP-CP-JSP (based on lower bound $\mathit{LowerBoundJPS}$). 
\cref{fig:dicho_nodes} compares the number of explored nodes needed to find and prove the optimality of JSP instances, while \cref{fig:dicho_times} compares the time needed by at least one model to do so.
The efficiency of filtering algorithms embedded into the \textsc{NoOverlap} global constraint significantly improves the lower bound, and thus reduces the number of explored nodes of almost all the instances.
The time spent improving the lower bound affects the execution time required to find and prove optimality in many instances. 
We notice that with the strong lower bound, it is possible to solve $30\times 10$ instances within the time limit.

\subsection{Question 4}
\begin{figure}[t] 
    \begin{minipage}{0.49\textwidth}
        \centering \includegraphics{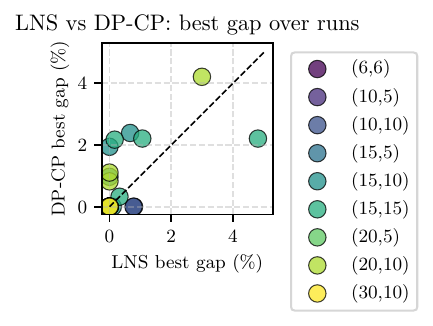}
        \caption{Best gap comparison. The gap is computed with respect to the best-known solutions (BKS) reported in the literature for the Lawrence and Taillard benchmarks.}
        \label{fig:lns_best_gap}
    \end{minipage}
    \hfill
    \begin{minipage}{0.49\textwidth}
        \centering        \includegraphics{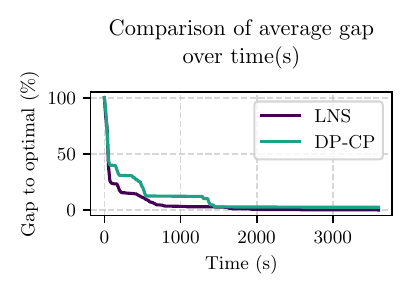}
        \caption{Average gap over time. The gap is computed with respect to the best-known solutions (BKS) reported in the literature for the Lawrence and Taillard benchmarks.}
        \label{fig:lns_gap_over_time}
    \end{minipage}
\end{figure}

The initial solution of our LNS is the first solution of the DP-CP model, and for each instance of the problem (JSP or OSP), we ran 10 random seeds to randomly select the fraction of precedences to keep.
For each run, 70\% of precedences are kept, and after 100 not fruitful restarts of the A*, the number of precedences to keep is reduced by 5\%. 
When the percentage is less than 20\%, the number of precedences is reset to its initial value, and the process is repeated until the time limit is reached.
\cref{fig:lns_best_gap} compares the best gap found at the end of the LNS with the gap at the end of the DP-CP model for each JSP instance.
We observe that within the time limit, the LNS gap is better than that of the DP-CP model for large instances. 
There are many instances for which the gap is 0\% for both searches.
\cref{fig:lns_gap_over_time} compares the average gap among the 10 runs of LNS on each instance with the gap of the DP-CP model over time.
We observe that the gap decreases more rapidly for the LNS than for the DP-CP model.
The LNS can find and prove the optimality of three instances of the Lawrence benchmarks \cite{lawrance1984resource} not proved by the hybridization model DP-CP.

On OSP, the LNS decreases the incumbent solution gap more rapidly than the DP-CP model.
The plots are similar to those of the JSP, and are reported in Section \ref{sec:results1} of the appendix.
Eight of the $10\times 10$ unsolved Taillard instances \cite{taillard1993benchmarks}  to optimality by the DP-CP model have been solved with the proof of optimality by the LNS. 
This illustrates the ability of our LNS to guide the search towards promising regions of the solution space.

\section{Conclusion}\label{sec:concl}

In this paper, we have proposed for the first time a hybridization of DP and CP for the partial shop scheduling problems like JSP or OSP.
The CP was used as a subroutine to leverage global constraint propagation, and the DP serves as a primary search framework.
The DP formulation uses the CP during the transition phase to reduce the branching factor of the search, and the CP is also used to compute the lower bound, which is useful for the ACS used to explore the search space of the DP.
The proposed model always reduces the explored nodes and has solved more instances than the CP model, sometimes with additional time.
It was able to quickly improve a solution by successively exploring a promising part of the search tree in a large neighborhood search approach.
The hybridization of DP and CP seems a promising research direction for solving other combinatorial optimization problems.

\bibliography{references}

\appendix
\section{Appendix}\label{sec:appendix}
In this section, remaining results are reported.
\subsection{Question 2}\label{sec:results} 
\begin{figure}[!ht] 
    \begin{minipage}{0.49\textwidth}
        \centering \includegraphics{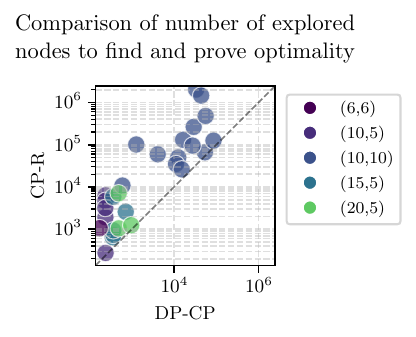}
        \caption{Nodes to optimality comparison}
        \label{fig:nodes_Rank}
    \end{minipage}
    \hfill
    \begin{minipage}{0.49\textwidth}
        \centering        \includegraphics{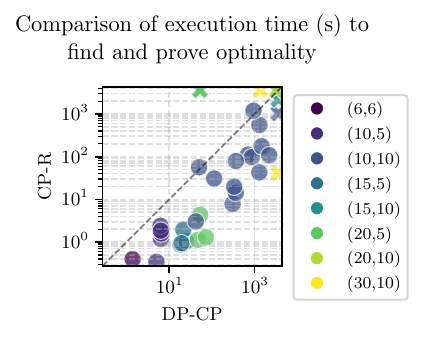}
        \caption{Time to optimality comparison}
        \label{fig:time_rank}
    \end{minipage}
\end{figure}

The second CP model, where the \textit{Rank} strategy search is compared to our hybridization model DP-CP on JSP instances.
\cref{fig:nodes_Rank} compares the number of nodes explored needed to find and prove the optimality of approaches, while \cref{fig:time_rank}  compares the execution time in seconds to find and prove the optimality on instances where at least one finds and proves the optimality.
Despite the systematic reduction of the number of explored nodes on almost all common solved instances, the running time is generally in favor of the rank CP model.

\begin{figure}[!ht]
    
    \centering        
    \includegraphics{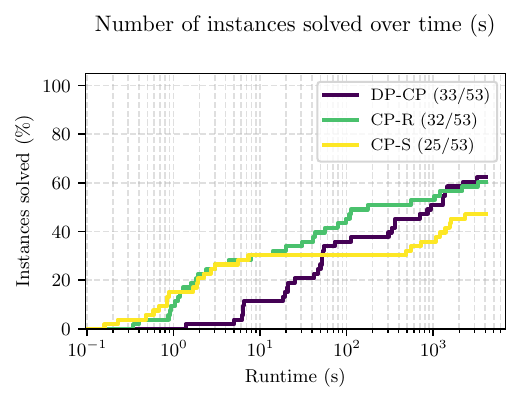}
    \caption{Solved instances over time (JSP)}
    \label{fig:cpvsDPCPall}
    
\end{figure}

\cref{fig:cpvsDPCPall} compares the number of JSP instances solved over time for the CP models against the hybridization DP-CP model.
For JSP instances solved with less than 100 seconds, CP approaches are superior to the DP-CP model, and the observation changes on hard instances where DP-CP solves more instances than CP-S.
DP-CP and CP-S are similar in the branching strategy since both branch on time, while CP-R branches on precedences, like in \cite{GrimesH10}.
DP-CP model solves more instances than the CP models, and the difference is more pronounced compared to CP-S.

\subsection{Question 4}\label{sec:results1} 
\begin{figure}[!ht] 
    \begin{minipage}{0.49\textwidth}
        \centering \includegraphics{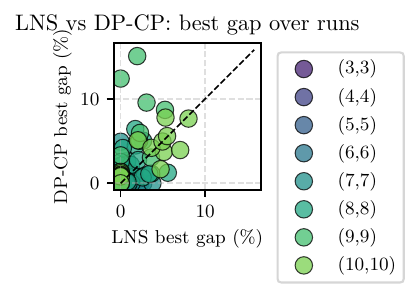}
        \caption{Best gap comparison over OSP}
        \label{fig:lns_best_gap1}
    \end{minipage}
    \hfill
    \begin{minipage}{0.49\textwidth}
        \centering        \includegraphics{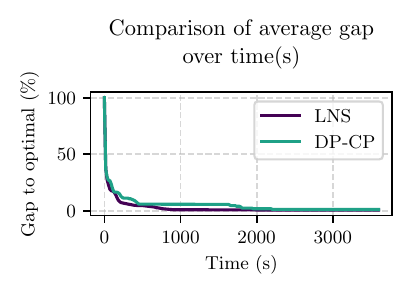}
        \caption{Average gap over timeover OSP}
        \label{fig:lns_gap_over_time1}
    \end{minipage}
\end{figure}

On OSP, the LNS decreases the incumbent solution gap more rapidly than the DP-CP model.
\cref{fig:lns_best_gap1} compares the best gap found at the end of the LNS with the gap at the end of the DP-CP model for each OSP instance.
We observe that within the time limit, the LNS gap is better than that of the DP-CP model for more instances. 
There are many instances for which the gap is 0\% for both approaches.
\cref{fig:lns_gap_over_time1} compares the average gap among the 10 runs of LNS on each instance with the gap of the DP-CP model over time.
We observe that the gap decreases more rapidly for the LNS model than for the DP-CP model.
The LNS can find and prove the optimality of eight of the $10\times 10$ unsolved Taillard instances \cite{taillard1993benchmarks} using the DP-CP model.
The instances $tai\_10\times10\_1$, $tai\_10\times10\_2$, $tai\_10\times10\_3$, $tai\_10\times10\_4$, $tai\_10\times10\_5$, $tai\_10\times10\_7$, $tai\_10\times10\_8$, and $tai\_10\times10\_10$ are those solved to optimality by the LNS.

\end{document}